\pdfoutput=1

\documentclass[11pt]{article}

\usepackage[final]{acl}

\usepackage{times}
\usepackage{latexsym}

\usepackage[T1]{fontenc}

\usepackage[utf8]{inputenc}

\usepackage{microtype}

\usepackage{inconsolata}

\usepackage{kotex}
\usepackage{color}
\usepackage{makecell}
\usepackage{xurl}
\usepackage{graphicx}
\usepackage{multirow}
\usepackage{graphicx}
\usepackage{subcaption}
\usepackage{booktabs}
\usepackage{xcolor, soul}
\usepackage{amssymb}
\usepackage{bbm}

\title{Assessing News Thumbnail Representativeness: \\Counterfactual text can enhance the cross-modal matching ability}

\author{
Yejun Yoon\textsuperscript{$\dagger$}~~~~~~Seunghyun Yoon\textsuperscript{$\ddagger$}~~~~~~Kunwoo Park\textsuperscript{$\dagger$$\star$}\\
\textsuperscript{$\dagger$}Department of Intelligent Semiconductors, Soongsil University\\
\textsuperscript{$\ddagger$}Adobe Research, USA\\
\textsuperscript{$\star$}School of AI Convergence, Soongsil University\\
\texttt{yeayen789@gmail.com}, \texttt{syoon@adobe.com}, \texttt{kunwoo.park@ssu.ac.kr}
}

\begin{document}
\maketitle
\begin{abstract}
This paper addresses the critical challenge of assessing the representativeness of news thumbnail images, which often serve as the first visual engagement for readers when an article is disseminated on social media. We focus on whether a news image represents the actors discussed in the news text. To serve the challenge, we introduce \textsc{NewsTT}, a manually annotated dataset of 1000 news thumbnail images and text pairs. We found that the pretrained vision and language models, such as BLIP-2, struggle with this task. Since news subjects frequently involve named entities or proper nouns, the pretrained models could have a limited capability to match news actors' visual and textual appearances. We hypothesize that learning to contrast news text with its counterfactual, of which named entities are replaced, can enhance the cross-modal matching ability of vision and language models. We propose \textsc{CFT-CLIP}, a contrastive learning framework that updates vision and language bi-encoders according to the hypothesis. We found that our simple method can boost the performance for assessing news thumbnail representativeness, supporting our assumption. Code and data can be accessed at \url{https://github.com/ssu-humane/news-images-acl24}.
\end{abstract}

\section{Introduction}

\begin{figure}[t]
\includegraphics[width=0.99\linewidth]{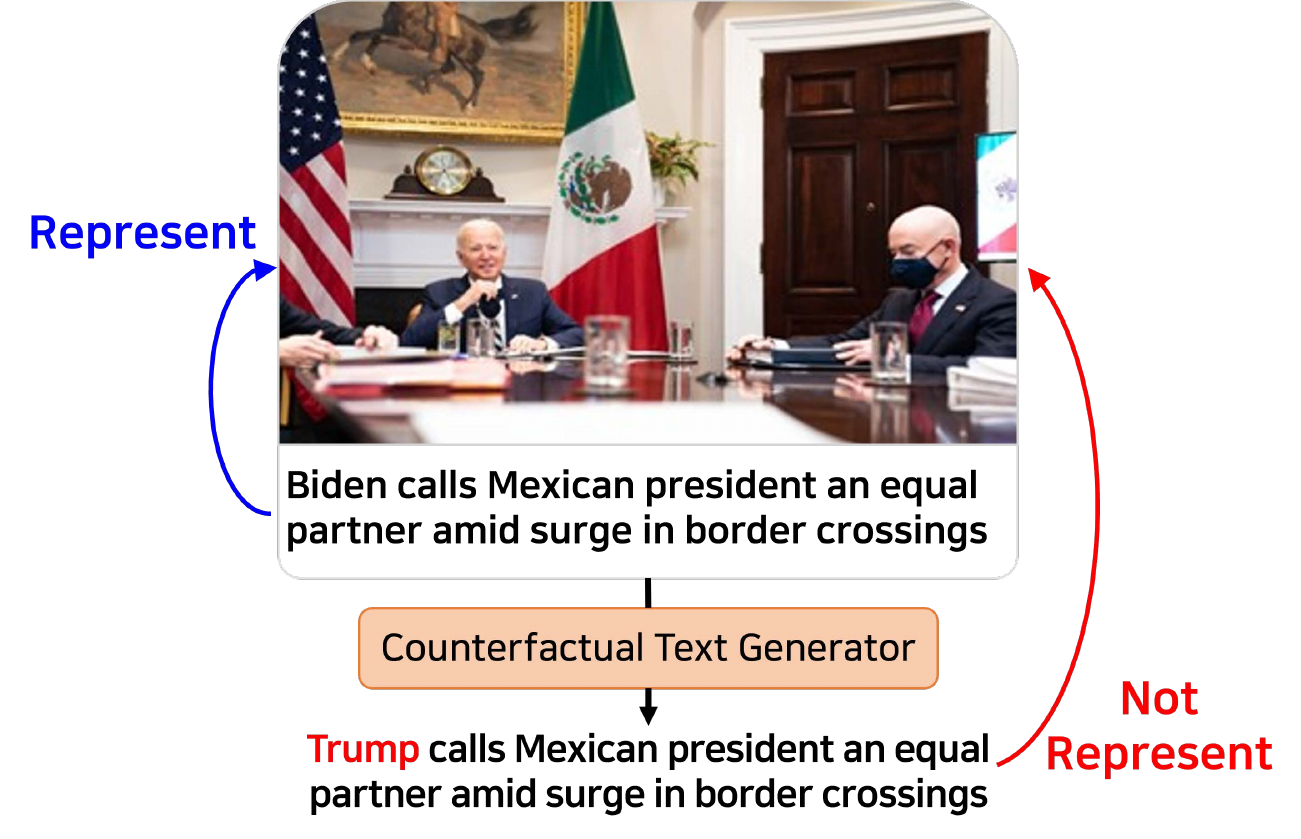}
\caption{An illustration of the key idea of the proposed method. To assess whether a news thumbnail image represents the body text, the method generates counterfactual text to be used as negative samples for contrastive updates.}
\label{fig:thumbnail}
\end{figure}

This study investigates the representativeness of news thumbnails, which are images displayed as previews of news articles. Visual content often carries a more persuasive impact and leaves a long-lasting impression compared to text~\cite{joffe2008power,newman2012nonprobative,seo2020meta}. Consequently, the misrepresentation in news thumbnails can cause more critical consequences. Despite its importance, there have been a few studies addressing the representativeness of news images~\cite{hessel2021clipscore,choi2022does}. Achieving the goal requires a model to understand the cross-modal relationship between visual and textual content. 

Researchers in Natural Language Processing and Computer Vision communities have explored multimodal technologies by addressing general vision and language (V+L) tasks, including visual question answering~\cite{antol2015vqa}, image-text retrieval~\cite{cao2022image}, and visual entailment~\cite{xie2019visual}. Recent approaches have leveraged contrastive learning techniques for representing distinct modalities in the same vector space. CLIP~\cite{radford2021learning} stands out as one of the pioneering methods that achieve remarkable performance, aimed at increasing the similarity between image-text pairs (referred to as \textit{positive} samples) compared to non-paired image and text (referred to as \textit{negative} samples). While contrastive learning was not a new concept~\cite{oord2018representation}, CLIP outperformed preexisting methods by utilizing a vast collection of web-based image and text data for pretraining. Recent advancements have further enhanced the performance by refining learning objectives~\cite{lisupervision}, incorporating cross-attention layers into the model architecture~\cite{li2021align}, and augmenting the training data with high-quality samples~\cite{li2022blip,Hao_2023_WACV}.

Despite the progress made in multimodal methods for general vision-and-language benchmarks, we hypothesize that evaluating the representativeness of thumbnails remains difficult, even for the most advanced models, due to two primary challenges. \textbf{(1) Preference for general descriptions in popular benchmarks}: General V+L datasets, such as Conceptual Captions~\cite{sharma-etal-2018-conceptual} or COCO-Caption~\cite{chen2015microsoft}, were primarily designed to enhance a model's ability to comprehend general behaviors or scenes. Consequently, these datasets deliberately avoid specific descriptions about entities (e.g., ``\ul{The man} at bat readies to swing at the pitch''). In contrast, news events typically revolve around political actors or organizations, leading to news texts that frequently reference named entities~\cite{luo-etal-2021-newsclippings}. \textbf{(2) Lack of dataset labeled according to an objective definition}: Representativeness is an inherently abstract and subjective concept, where even humans may disagree with each other. Developing an objective definition is imperative to obtain high-quality labels that different human annotators can agree on and facilitate the development of proper technologies. While there was a human-labeled dataset~\cite{choi2022does}, they asked for subjective opinions of human annotators without providing objective definitions grounded by journalism principles.

To address (1) and (2), borrowing the principle of five Ws in journalism~\cite{apstylebook}, we introduce the evaluation dataset of \textsc{news} \textsc{t}humbnails and \textsc{t}ext pairs, \textsc{NewsTT}. The annotators were instructed to label the 1000 pairs through the lens of \textit{Who}, on whether an image represents news actors. Using the high-quality labels reflecting the objective definition of representativeness, we evaluated the zero-shot ability of vision and language models to understand news actors' visual portrayals. We found that CLIP performed better at the task than more recent methods, such as BLIP-2, suggesting the efficiencies of its dual encoder architecture and contrastive objective. To improve its matching ability, this study introduces \textsc{CFT-CLIP}, a \textsc{c}ounter\textsc{f}actual \textsc{t}ext-guided \textsc{c}ontrastive \textsc{l}anguage-\textsc{i}mage \textsc{p}retraining framework. As illustrated in Figure~\ref{fig:thumbnail}, the key idea of the proposed framework is to generate counterfactual news text where the actor of a news event is changed to be used as hard negatives for contrastive learning. In the news text example, the actor is identified as `Biden,' and its corresponding token is replaced with `Trump' by the counterfactual text generator. Thus, the generated sentence no longer represents the key actor of the news event depicted in the thumbnail image. Our research hypothesis is that \emph{learning to contrast the generated counterfactual text with the original news text will enhance the ability of vision language models to assess the representativeness of a thumbnail image}.

We summarize the contributions of this study. 
\begin{enumerate}
    \item We address the problem of assessing news thumbnail representativeness by determining whether a given news image depicts the actors of the news events (Who) according to the principle of 5Ws. 
    \item To serve the task, we introduce \textsc{NewsTT}, a manually annotated dataset of 1,000 news thumbnails and text pairs with high-quality labels. We found that general vision language models struggle with it.
    \item We propose \textsc{CFT-CLIP}, a counterfactual text-guided contrastive language-image pretraining framework. We found that our simple method could outperform larger pretrained and domain-adapted models, supporting the research hypothesis.
\end{enumerate}

\section{Related Works}
\subsection{Vision language contrastive pretraining}
Researchers have worked on pretraining methods to tackle V+L tasks such as image-text retrieval. \citet{lu2019vilbert} extended BERT to a multi-modal model by co-attentional transformer layers. Another study proposed to learn a universal image and text representation by four pre-training tasks, including masked language modeling and image-text matching~\cite{chen2020uniter}. Researchers proposed CLIP, a training scheme of visual representation by learning directly from paired text using contrastive loss~\cite{radford2021learning}. Being trained on a web-scaled dataset, the transformer-based visual representation achieved state-of-the-art results in zero-shot tasks. Another study showed its potential as a reference-free image-captioning metric~\cite{hessel2021clipscore}. Other researchers presented ALIGN, a bi-encoder trained by a contrastive loss similar to CLIP~\cite{jia2021scaling}. While we used CLIP encoders as the target backbone in this study, our contrastive learning framework can be applied to a more recent vision-and-language model that incorporates the image-text alignment, such as BLIP-2~\cite{li2023blip}. 

Research has underscored the significance of incorporating hard negative samples into the contrastive objective~\cite{nishikawa2022ease,robinson2020contrastive}. In response, we introduce a method to generate hard negative text samples using a masked language model. Although a handful of studies have explored the use of masked language models for generating hard negatives~\cite{li-etal-2021-unimo,yao2022contrastive}, our approach stands apart due to its distinct token selection and prediction procedures in the step of counterfactual text generation. We compared its effectiveness against the autoregressive generation of GPT-3.5 Turbo in the ablation experiments.

\subsection{Multimodal misinformation}

This study falls into the broad category of research on multimodal misinformation. Previous research mainly worked on fact verification involving multimodal claims or evidence~\cite{zlatkova-etal-2019-fact,shu2020fakenewsnet,yao2023end,rani2023factify}. In contrast, this study examines whether news thumbnails and text present the same actor. A relevant problem is image repurposing~\cite{luo-etal-2021-newsclippings,abdelnabi2022open}, where a \textit{real} image is used out-of-context to create realistically looking misinformation. To tackle the problem, \citet{luo-etal-2021-newsclippings} proposed NewsCLIPPings, a dataset of out-of-context image detection in the news context. They created the mismatched image-text pairs by swapping the matches of the collected news image and text pairs. Similarly, \citet{mishra2022factify} created a dataset of multimodal fact verification, of which false claims were obtained by manipulating the pairs. Unlike the previous research on out-of-context image detection, 
this study aims to focus on whether news subjects are represented in the thumbnail image. We also created a manually annotated evaluation dataset, contrasting with previous datasets based on synthetic labels~\cite{luo-etal-2021-newsclippings,mishra2022factify}. Our target task is related to but differs from the manipulated image detection~\cite{huh2018fighting,rossler2019faceforensics}, known as deepfakes or cheapfakes, which aimed to detect a \textit{fake} image generated by AI technologies or simple tools.

\section{Target Problem: Assessing News Thumbnail Representativeness}
\label{sec:problem}

We aim to address the problem of automatically evaluating the representativeness of a news thumbnail image for its article. The task is critical due to the two reasons. First, the sharing of news online is abundant~\cite{park2021present,lee2012news,hermida2012share}. Second, the thumbnail often serves as the initial and sole visual interaction for social media users when a news link is shared. Visual content is perceived as more credible and leaves a long-lasting impression than text~\cite{joffe2008power,seo2020meta,newman2020non}. Given these considerations, thumbnail images that do not accurately represent their content can misguide the reader's understanding in online environments. 

We formally define the problem as a binary classification. Given a news thumbnail $I$ and its news text $T$, we aim to develop a classifier $f_{\theta}(I, T)$ predicting a binary label $L$ on predicting whether $I$ represents $T$. We assign 1 to $L$ if $I$ represents $T$; otherwise, 0. $I$ is deemed representative of $T$ if $I$ portrays at least one of the actors of a news event, which can be identified from $T$. This study aims to develop a vision language model that predicts $L$ from a pair of $I$ and $T$. We assume the task in a zero-shot setting, where the labeled data is limited. The reference news text $T$ can be either a title or a summary.

\begin{figure*}
    \centering
    \includegraphics[width=0.8\textwidth]{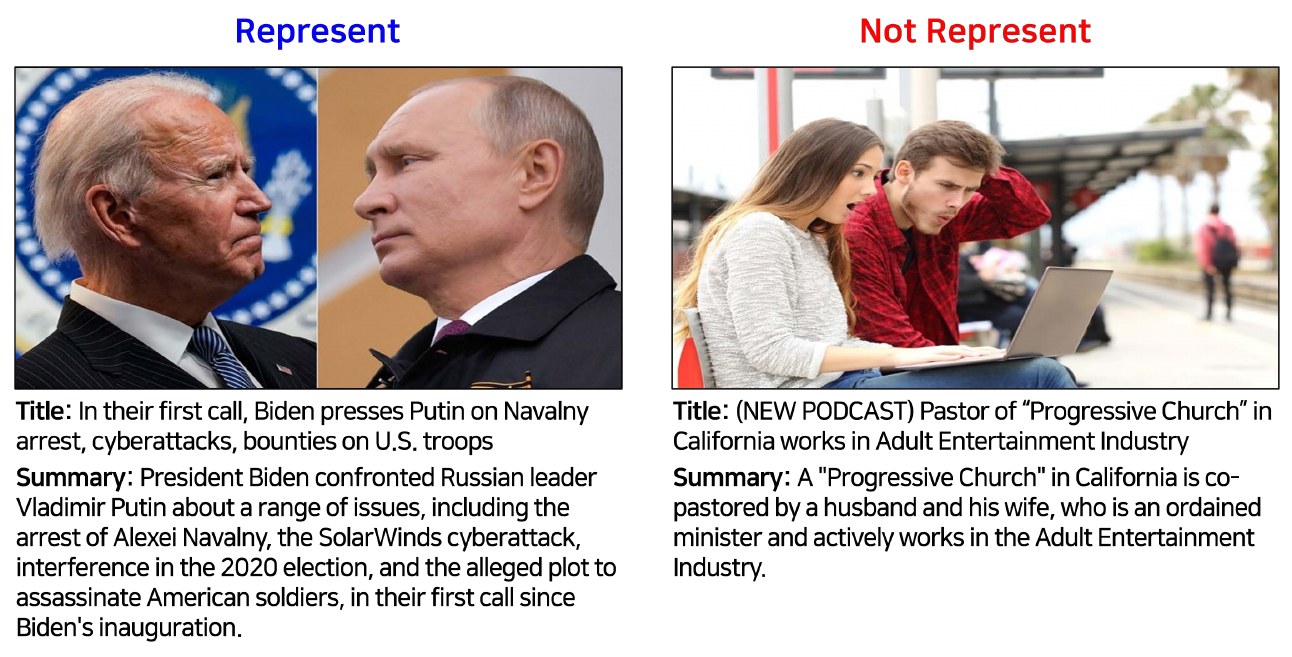}
    \caption[Annotation label example.]%
    {Labeled data examples.}
    \label{fig:annotation_label_example}
\end{figure*}

\section{\textsc{NewsTT}: A Dataset of Thumbnail Representativeness for News Text}

\subsection{Raw data collection} 

NELA-GT-2021 is a headline-oriented\footnote{They put the body text whose random tokens are masked.} news corpus that encompasses nearly all news articles published by 367 news sources throughout the year 2021~\citep{gruppi2022nela}. Since the dataset does not provide web links to thumbnail images, we conducted a supplementary data collection step. By referencing the URLs of news articles available in the dataset, we parsed the HTML document and extracted the thumbnail image URL from the meta tag with the og:image attribute. As a result, we obtained 442,741 pairs of news headlines and thumbnail images, which were sourced from 81 different news media outlets. According to the media ratings provided by Media Bias/Fact Check (MBFC)\footnote{\url{https://mediabiasfactcheck.com/}}, the datasets were well-balanced for the political orientation and trustworthiness ratings. Subsequently, we partitioned this unlabeled dataset into 427,741/5,000/10,000 pairs for train/validation/test purposes while keeping the distribution of the number of articles over different media sources. The validation split was reserved for hyperparameter optimization, and the test split served as the source for labeled data construction. 

\subsection{Data annotation}
\label{sec:annotation}

We describe the process for building the annotation guideline. Following a previous study~\cite{choi2022does}, we conducted the pilot labeling task given an abstract definition of a representative thumbnail, which is ``\textit{an image that visually conveys the news event that can be identified from the accompanying text}''. There were often disagreements in the resulting labels likely due to the subjective nature of definition. Following internal discussions and consultation with a journalism expert with a Ph.D. in Communication, we focused on the key dimension of news events corresponding to the five Ws~\cite{apstylebook}: on whether the image represents news subjects, i.e., \textit{Who}.

For the annotation process, we sampled 1,000 news articles from the pool of 10,000 instances of the test split. Relying on the MBFC ratings, we sampled 500 articles from the trustworthy outlets, labeled as \textit{mixed} or \textit{high} for the trustworthiness rating, and the remaining 500 from the fake news sources, labeled as \textit{low}. We manually collected the body text for the 1000 articles and ran the OpenAI API for obtaining the summary text by GPT-3.5 Turbo\footnote{The used prompt is in Appendix~\ref{appendix:chatgpt}.}. We hired three English-speaking students who frequently read news online from one of the authors' institutions. Given a news thumbnail, title, and summary text, the annotators were asked to answer whether the news actors identified in the news text are visually portrayed in the thumbnail image. To facilitate the annotation process, we instructed them to identify the actors of the news event by referring to the text and, in turn, to find visually depicted ones. Detailed annotation guidelines can be found in Appendix~\ref{sec:guideline}. The annotation results demonstrated a high inter-annotator agreement rate, with a Krippendorff's alpha of 0.705, ensuring the quality and reliability of the labels. 

\begin{figure*}[t!]
    \centering
    \includegraphics[width=0.8\textwidth]{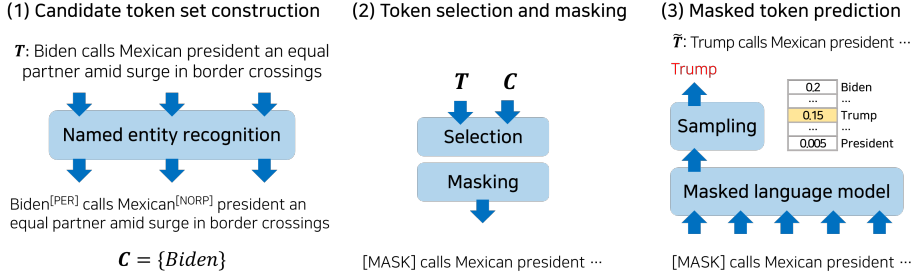}
    \caption
    {An illustration of the counterfactual text generation process by CFT-CLIP. We select named entity tokens in an original text that could indicate news subjects. A masked language model generates a counterfactual text by predicting new tokens for the selected entity tokens.}
    \label{fig:hard_negative_generation_example}
\end{figure*}

\begin{table}[t]
\small
\centering
\begin{tabular}{ccccc}
\toprule
& \multicolumn{2}{c}{Represent (1)} & \multicolumn{2}{c}{Not represent (0)} \\\midrule
  & Title & Summary & Title & Summary \\\midrule
Words & 14.9 & 39.1 & 14.7 & 39 \\
Nouns & 3.45 & 8.63 & 3.75 & 9.7 \\
Verbs & 1.81 & 4.06 & 1.68 & 4.09 \\
Adjectives & 0.93 & 2.39 & 1.19 & 2.73 \\
Named entities & 2 & 3.41 & 1.64 & 2.76\\\bottomrule
\end{tabular}
\caption{Mean counts of words, part-of-speech units, and named entities measured on the labeled text.}
\label{tab:data_analysis}
\end{table}

\subsection{Data analysis}

\textsc{NewsTT} consists of 817 representative and 183 non-representative image-text pairs. Table~\ref{tab:data_analysis} presents the descriptive characteristics of the dataset for each class. While the samples in the two classes are similarly long, the news text of label 1 tends to include more named entities. Figure~\ref{fig:annotation_label_example} presents an example for each class. The label 1 example shows the image portraying Biden and Putin, who are the actors of a news event. The thumbnail image on the right does not show the news actor but arouses the potential reader's interest by showing the surprised faces.

\section{Methods}
To tackle the task of assessing thumbnail representativeness in a zero-shot setting where no labeled data is available for training, we use a thresholding classifier based on the cross-modal similarity,
\begin{equation}\label{eq:classifier}
    f(I, T)=\mathbbm{1}(sim(\mathbf{v}_{I},\mathbf{v}_{T}) > \tau),
\end{equation}
where $\tau$ is a thresholding hyperparameter, $sim(\cdot)$ is cosine similarity, $\mathbbm{1}(\cdot)$ is an indicator function, and $\mathbf{v}_{I}$ and $\mathbf{v}_{T}$ indicate the feature embedding of $I$ and $T$, respectively. Assuming the parity prior, we set $\tau$ as the median of $sim(\cdot)$ on the validation set.

\subsection{Background: CLIP}

CLIP is a vision and language bi-encoder that represents each modality in a fixed-dimensional vector space~\cite{radford2021learning}. The model is based on a bi-encoder architecture, $f_{image}(\cdot)$ and $f_{text}(\cdot)$. For an input image $I$ and text $T$, the two encoders return the image embedding $\mathbf{v}_I$ and the text embedding $\mathbf{v}_T$, respectively. The parameters were trained via contrastive learning on a large web collection of image-text pairs. Since $\mathbf{v}_I$ and $\mathbf{v}_T$ are in the same vector space, CLIP can serve as a backbone for the zero-shot thresholding classifier in Eq.~\ref{eq:classifier}.

\subsection{Proposed method: \textsc{CFT-CLIP}}
We propose CFT-CLIP, a counterfactual text-guided contrastive language-image pretraining method. The proposed framework aims to improve the vision and language bi-encoder by contrastive updates involving the counterfactual text generated from an input text. Figure~\ref{fig:hard_negative_generation_example} shows an example of generation. While the original news pertains to an event involving Biden's summit with the Mexican president, the generated sentence implies a counterfactual event centered on Trump by substituting the token `Biden'. This substitution is accomplished via the prediction of a masked language model targeting the named entity token. Based on the news characteristics on the prevalent coverage of named entities~\cite{park2021blames,muller2020multimodal}, we hypothesize that contrasting the original news text with its counterfactual would make the vision and language bi-encoder to obtain a more effective representation for understanding a thumbnail image's representativeness.

\subsubsection{Counterfactual text generation}

Given a pair of image $I$ and text $T$, we aim to generate a counterfactual news text $\tilde{T}$, which is semantically distinct from the anchor image $I$ (and the original text $T$). 

\noindent\textbf{(1) Candidate token set construction}: The first step aims to select tokens that likely refer to the actor of a news event. A part-of-speech tagger and named entity recognizer are applied to $T$, which returns a candidate token set $C$. We tested several strategies targeting named entity categories that are frequently used in news articles, such as person, organization, and GPE\footnote{Countries, cities, states}.

\noindent\textbf{(2) Token selection and masking}: We select tokens from $C$ and mask them in the original text $T$ using the special token \textsc{[MASK]}. To preserve the original context of $T$, we ensure that no more than 30\% of its total tokens are masked. If the number of target tokens selected from $C$ exceeds this threshold, we randomly select a subset of these tokens to meet the condition. Otherwise, all tokens in $C$ are masked in $T$. 

\noindent\textbf{(3) Masked token prediction}: Using a masked language model, e.g., BERT~\cite{devlin-etal-2019-bert}, we predict new tokens for the masked positions in $T$. To avoid reconstructing the original text, we divide the logit scores of the output vector by a temperature parameter and repeatedly sample from the softmax-normalized distribution until a token different from the original is generated. This step produces $\tilde{T}$, which represents a synthetic event wherein the subject is modified from the original news event, yet it preserves the overarching topic of $T$. Thus, $\tilde{T}$ can be deemed a hard negative sample in contrastive learning, which diverges from $I$ but remains more closely aligned with $T$ than a random text or in-batch negatives. 

In Section~\ref{sec:abl}, we present the ablation experiments to show the effectiveness of the proposed counterfactual text generation. Additionally, we demonstrate several counterfactual text examples generated by the proposed method in Appendix~\ref{sec:app:counter}.

\subsubsection{Training objective}

We train an image and text bi-encoder model to minimize
\begin{equation}\label{eq2}
    -log\frac{e^{sim(\mathbf{v}_{I_i},\mathbf{v}_{T_i})/\tau}}{\sum_{j=1}^{N}\{e^{sim(\mathbf{v}_{I_i},\mathbf{v}_{T_j})/\tau}+e^{sim(\mathbf{v}_{I_i},\mathbf{v}_{\tilde{T}_j})/\tau}\}},
\end{equation}
where $\mathbf{v}_{I_i}$ is the feature embedding of the vision encoder for $I_i$, $\mathbf{v}_{T_i}$ is the feature embedding of the text encoder for $T_i$, $sim(\cdot)$ is the dot product, and $N$ is the mini-batch size.

Minimizing Eq.~\ref{eq2} aligns the text representation with the image representation by increasing the similarity of positive pair $(I_i, T_i)$ compared to those of negative pairs $(I_i, T_j)$ and $(I_i, \tilde{T}_i)$. Since $\tilde{T}$ represents a distinct news event of which actors were replaced from $T$, the objective would help a bi-encoder model learn to capture whether the thumbnail image represents the news actors that can be identified from $T$.

\subsubsection{Model architecture}

\begin{figure}[t]
    \centering
    \includegraphics[width=0.95\linewidth]{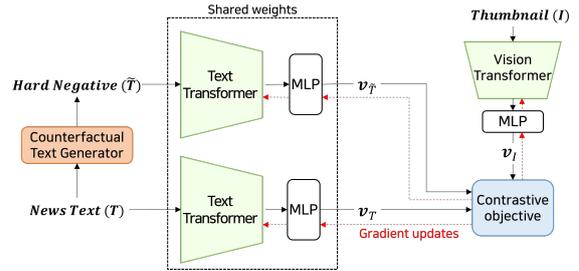}
    \caption{Neural architecture of CFT-CLIP}
    \label{fig:architecture}
\end{figure}

Figure~\ref{fig:architecture} depicts the neural architecture used for the proposed method. We initialized the model by adopting a pretrained CLIP checkpoint\footnote{\url{https://huggingface.co/openai/clip-vit-large-patch14}}. The image encoder $f_{image}$ is the ViT-L/14 vision transformer (24 layers), and the text encoder $f_{text}$ is a causal text transformer (12 layers). The image and text encoder are followed by an MLP pooler mapping the input to a 768-dimensional vector, respectively. $T$ and $\tilde{T}$ are fed into the shared-weight text encoder, separately. According to the hyperparameter optimization experiment, we froze 11 out of 12 text transformer layers during training. All vision layers were updated. 

\subsubsection{Pretraining corpus}

We used two pretraining corpora. First, we used the training split of \textsf{NELA} unlabeled data. Inspired by a prior study that replicated the CLIP pretraining based on a web data collection~\cite{schuhmannlaion}, we applied the data sanitization process to the training split of unlabeled data by filtering only those with a cosine similarity of 0.28 or higher over CLIP embeddings. The resulting 105,737 pairs for the unlabeled training have a high likelihood of a representative thumbnail ($L=1$). Since the body text in the NELA corpus is incomplete, we relied on the news title as the reference text for pretraining. Second, inspired by the recent research on the importance of high-quality samples for pretraining~\cite{zhou2023lima,schuhmannlaion}, we used a high-quality news corpus published by the \textsf{BBC}, a UK-based news source rated as trustworthy by MBFC~\cite{verma-etal-2023-large}. This corpus comprises 196,538 pairs, each containing a news title, an editor-written summary, and a thumbnail image. The corpus encompasses news articles released through the official website from 2009 to 2021. We applied the sanitation process to the dataset in the same manner as with the NELA corpus, leaving 58,210 samples for contrastive training. We ensured no overlap between the BBC training set and the NELA validation/test splits. Since NewsTT originates from the NELA corpus, we used the NELA validation split for hyperparameter optimization. In the experiments using the BBC corpus, we tested two variants using the news title and human-written summary as reference text.

\section{Evaluation}

We conducted evaluation experiments to understand the effects of the proposed method and test the research hypothesis. In Section \ref{sec:main_results}, we examined how the pretrained and proposed vision language models perform for the target task. In ablation experiments (Section \ref{sec:abl}), we analyzed the effects of each module of the proposed method in more detail. Lastly, in Section \ref{sec:qualitative}, we conducted qualitative analyses to identify the error patterns where the proposed method fails.

\subsection{Experimental setups}

As baselines, we measured the ability of pretrained vision language models. In addition to \textsf{CLIP}, we used \textsf{BLIP} and \textsf{BLIP-2}, which are a family of vision language encoders that is based on bootstrapped training~\cite{li2022blip,li2023blip}. They achieved state-of-the-art performance in the vision-language benchmarks. \textsf{BLIP-2$+$SBERT} is a pipelined approach that integrates BLIP-2 with SentenceBERT. By obtaining a caption text from an input image $I$ by BLIP-2, we assessed its semantic similarity to the reference text using SentenceBERT. To investigate the effects of domain-adaptive pretraining, we also included \textsf{CLIPAdapt} as a baseline model, which was trained by using the CLIP objective on the training corpus. For inference, we used the summary text as the reference text $T$ according to the results of an ablation experiment (Table~\ref{abl:infer_text}).

To evaluate the baseline and proposed methods, we employed the f1 score and the Spearman rank correlation coefficient to evaluate binary prediction and cosine similarity scores, respectively. All experiments were conducted using five different random seeds, and we reported the average performance along with the standard error. The t-test was used for estimating statistical significance. We used the SpaCy pipeline for named entity recognition. A BERT-base checkpoint was used for the masked language prediction. All checkpoints used for experiments and implementation details can be found in Appendix~\ref{app:config}. We optimized all hyperparameters using the NELA validation split.

\begin{table}[t]
\small
\centering
\begin{tabular}{cccc}
\toprule
Model & F1 & Spearman\\\midrule
CFT-CLIP & \textbf{0.815}$\pm$0.003 & \textbf{0.491}$\pm$0.005\\
CLIPAdapt & 0.767$\pm$0.006 & 0.459$\pm$0.004 \\\midrule
CLIP & 0.763 & 0.409 \\
BLIP & 0.737 & 0.408 \\
BLIP-2 & 0.707 & 0.415 \\
BLIP-2 $+$ SBERT & 0.694 & 0.341 \\\bottomrule
\end{tabular}
\caption{Model comparison results}
\label{eval}
\end{table}

\subsection{Main results}
\label{sec:main_results}

Table~\ref{eval} presents the evaluation results of the baseline and proposed methods. Here, we reported the performance of CFT-CLIP and CLIPAdapt models that were pretrained using the BBC dataset with the summary text. For CFT-CLIP, we used the model targeting person-labeled entity tokens. The decisions were based on the ablation experiments on the effect of pretraining corpus (Table~\ref{abl:pretrain_corpus}) and counterfactual text generation (Table~\ref{abl:targeted_token}), respectively. We made three observations. First, among the pre-trained models (the bottom four rows), CLIP achieved the best f1 of 0.763, and BLIP-2 was the best by the Spearman coefficient with 0.415. Given that BLIP-2 (1.17B) is 2.74 times larger than CLIP (427M), this observation suggests the effectiveness of CLIP's bi-encoder architecture for the target task. Second, CLIPAdapt outperformed all the pretrained models. This suggests that domain-adapted continued pretraining can improve the performance for the target task, congruent with the finding of a previous study~\cite{gururangan-etal-2020-dont}. Third, CFT-CLIP outperformed all baseline methods with the f1 of 0.815 and the Spearman coefficient of 0.491 (\textit{p}$<$0.001). The performance gap with CLIPAdapt, the second-best method, is significant: 0.048 for f1 and 0.032 for the Spearman coefficient. The finding suggests that incorporating the counterfactual text in its contrastive objective as hard negatives can improve the cross-modal matching ability of the vision language bi-encoder, supporting the research hypothesis.

\subsection{Ablation experiments}
\label{sec:abl}

\begin{table}[t]
\small
\centering
\begin{tabular}{cccc}
\toprule
Target token & F1 & Spearman\\\midrule
Person  & \textbf{0.815}$\pm$0.003 & \textbf{0.491}$\pm$0.005 \\
Organization & 0.784$\pm$0.002 & 0.443$\pm$0.002 \\
GPE & 0.762$\pm$0.004 & 0.410$\pm$0.005 \\
All & 0.785$\pm$0.003 & 0.457$\pm$0.003 \\\midrule
Random (15\%) & 0.715$\pm$0.013 & 0.463$\pm$0.005 \\ 
Random (30\%) & 0.68$\pm$0.007 & 0.461$\pm$0.002 \\\bottomrule
\end{tabular}
\caption{Varying performance by token selection strategies in counterfactual text generation}
\label{abl:targeted_token}
\end{table}

\paragraph{What tokens should be targeted?} We evaluated the performance of CFT-CLIP variants aiming to replace different types of entity tokens in counterfactual text generation. In particular, we targeted the frequently used named entity tokens in the news text: person, organization, and GPE. We also tested the model that replaces all three token types, denoted All in the table. The top 4 rows in Table~\ref{abl:targeted_token} present the comparison results. We found that targeting person-labeled entity tokens achieved the best performance. This finding could be explained by the prevalent coverage of person entities, such as politicians, in news events~\cite{park2021blames,muller2020multimodal}. Additionally, we tested baseline strategies that replace tokens at random positions for being used for negative samples in contrastive learning, as investigated in previous studies~\cite{nishikawa2022ease,robinson2020contrastive}. We evaluated two variants that select 15\% and 30\% tokens, respectively, which were predicted by the same BERT backbone as the proposed method. The bottom two rows in Table~\ref{abl:targeted_token} show the model performance, achieving significantly worse f1 and Spearman coefficients than the CFT-CLIP variants (\textit{p}$<$0.001). According to the results, we targeted person-labeled tokens in the proposed method.

\paragraph{Is the masked LM necessary?} CFT-CLIP used a masked language model to generate counterfactual text for contrastive updates. An autoregressive large language model, such as GPT~\cite{radford2018improving}, could be used alternatively. To validate the idea, we ran the OpenAI API for generating counterfactual text for the BBC news summary by GPT 3.5-Turbo. The generated text was used for contrastive update, following the objective in Eq.~\ref{eq2}. The used prompt is available in Appendix~\ref{appendix:chatgpt}. As shown in Table~\ref{abl:gpt}, the GPT-based contrastive update was not as successful as the proposed method based on a masked language model. The proposed select-and-replace approach could generate a more suitable counterfactual sample to be used for hard negatives, rather than rewriting the whole sentence conditioned on the original text. 

We also tested a baseline approach that ablates a generation model for predicting masked tokens by randomly sampling person-labeled entities from the tokens from the training set. The simple method achieved an f1 of 0.77 and a Spearman coefficient of 0.455, respectively. This finding supports the effectiveness and necessity of counterfactual text generation in the proposed method.

\begin{table}[t]
\small
\centering
\begin{tabular}{cccc}
\toprule
Model & F1 & Spearman\\\midrule
CFT-CLIP  & \textbf{0.815}$\pm$0.003 & \textbf{0.491}$\pm$0.005 \\
GPT-based & 0.640$\pm$0.018 & 0.445$\pm$0.002 \\\bottomrule
\end{tabular}
\caption{Comparison with the model trained with GPT-based counterfactual text}
\label{abl:gpt}
\end{table}

\begin{table}[t]
\small
\centering
\begin{tabular}{cccc}
\toprule
Data & F1 & Spearman\\\midrule
BBC ($T$: summary) & \textbf{0.815}$\pm$0.003 & 0.491$\pm$0.005 \\
BBC ($T$: title) & 0.790$\pm$0.007 & \textbf{0.504}$\pm$0.001\\
NELA ($T$: title) & 0.772$\pm$0.003 & 0.448$\pm$0.002\\
\bottomrule
\end{tabular}
\caption{Varying performance by the pretraining corpus}
\label{abl:pretrain_corpus}
\end{table}

\paragraph{Data quality vs. quantity} Table~\ref{abl:pretrain_corpus} presents the performance of CFT-CLIP models with varying pretraining corpus. The BBC dataset resulted in superior model performance compared to the NELA dataset, following the same distribution of the labeled dataset. Given that the BBC is recognized for upholding high journalistic standards, this observation may indicate that leveraging a high-quality singular source is more beneficial than employing articles published by diverse news sources for contrastive pretraining. This finding is aligned with the recent research on large language models~\cite{zhou2023lima}, emphasizing that ensuring data quality is more important than merely increasing the size of the training corpus. In the experiments comparing title and editor-written summaries using BBC, we could not find a clear winner. While using the summary text led to the best f1 of 0.815, using the title achieved the best Spearman coefficient of 0.504. This suggests that both news headlines and summary text can serve as a useful proxy for news content, in line with established journalism principles~\cite{apstylebook}. Since f1 is a more proper metric for evaluating classification ability, we used BBC with the summary text for the other experiments. 

\begin{figure*}[t]
    \centering
    \includegraphics[width=.7\linewidth]{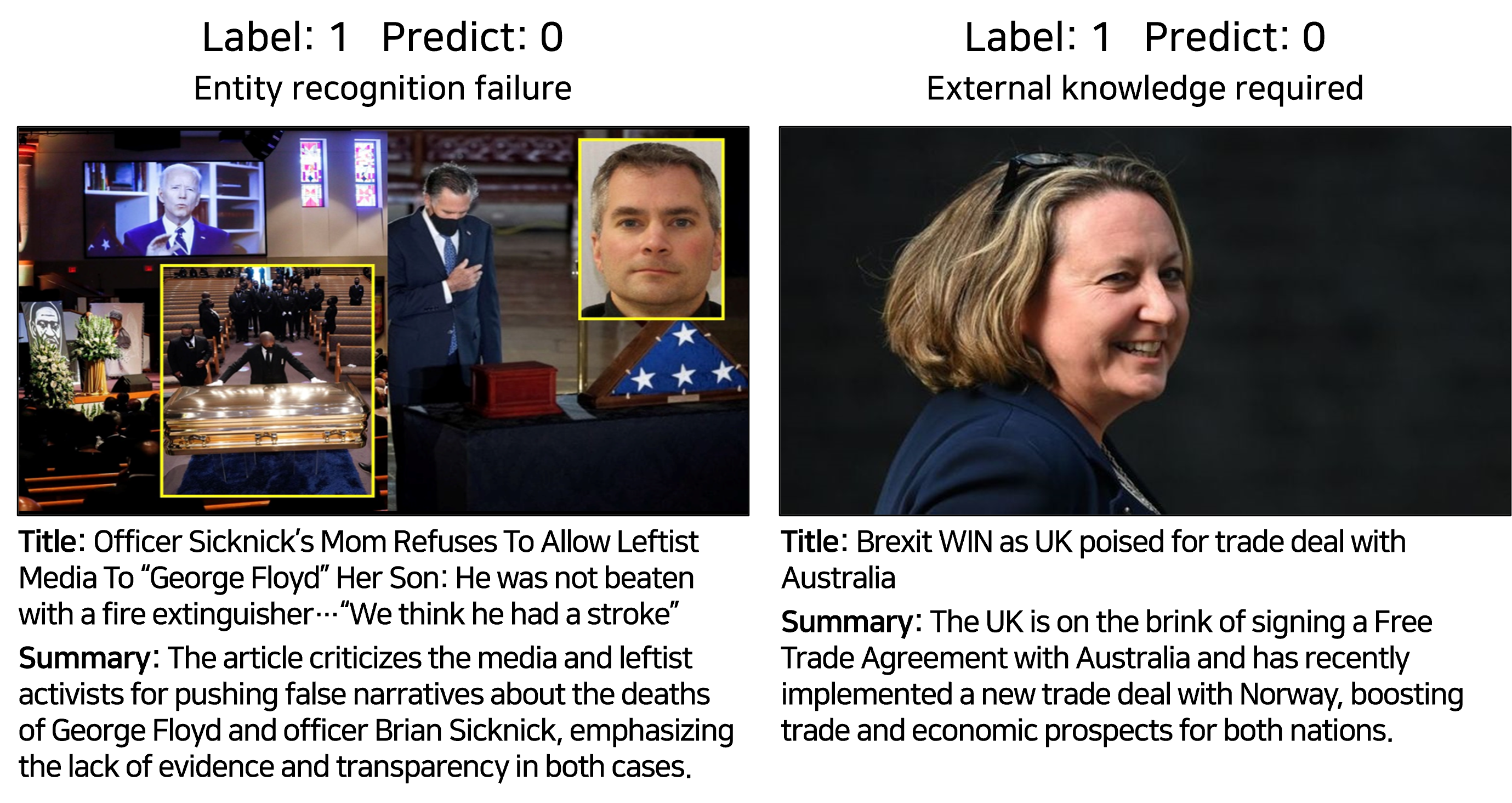}
    \caption{Error examples}
    \label{fig:qualitative_failed_example}
\end{figure*}

\subsection{Error analysis}
\label{sec:qualitative}

To identify the remaining challenges, we analyzed error categories for the sampled 100 error cases. The first author identified the initial category by thematic coding, which were improved by the iteration of discussion with the other authors and re-annotation. We observed four recurring categories: entity recognition failures (46\%), external knowledge required (14\%), deep textual understanding required (14\%), and deep visual understanding required (14\%). Figure~\ref{fig:qualitative_failed_example} presents two error examples. In the first example, which belongs to the entity recognition failure case, the model made a wrong prediction possibly due to the failed identification of Brian Sicknick, the officer involved in the George Floyd case. In the second example, while the image shows Anne-Marie Trevelyan, a UK minister, the model could not find the cross-modal link without referring to external knowledge.

\section{Discussion and Conclusion}

Automatically evaluating a news thumbnail image's representativeness is important for holding journalistic standards and building a trustworthy online environment. To address the important but underexplored problem in the research community, this study introduced NewsTT, a paired dataset of news thumbnail images and text with high-quality labels on whether the image represents the actor of the news event. We investigated the use of vision language models for the zero-shot assessment. Our proposed CFT-CLIP outperformed larger pretrained vision language models and domain-adapted methods. This supports the research hypothesis that counterfactual news text, of which named entities are replaced by a masked language model, could enhance the cross-modal matching ability by contrastive learning.

In ablation experiments, we found that using the counterfactual text generated by an autoregressive transformer language model, GPT-3.5 Turbo, as negative samples could not achieve a better outcome than CFT-CLIP. While this supports the effectiveness of the proposed method, we do not conclude that masked language models are clear winners for the counterfactual news generation over autoregressive language models. After manually examining the results, we observed that GPT could produce plausible counterfactual news. By contrast, the generation by a masked language model sometimes led to imperfect generations, including broken grammar (Table~\ref{tab:counterfactual_examples}). As shown in a recent study on text embedding~\cite{lee2024gecko}, an additional step might be required to ensure the quality of the generated text.  

This study has several future directions. First, according to the rule of 5Ws, this study focused on the \textit{Who} aspect of news thumbnails. Future research could extend its focus to cover other aspects of 5Ws, such as whether the image represents the subject's action, i.e., \textit{What}. Question-answering approaches could be investigated, as done in a recent study~\cite{rani2023factify}. Second, the proposed method could be improved to address the challenges identified in the error analysis (Section~\ref{sec:qualitative}). Future studies could investigate the use of multimodal language models, such as InstructBLIP~\cite{instructblip}, LLaVA~\cite{liu2024visual}, or GPT-4V~\cite{gpt4v}, to allow for handling the errors involving \emph{deep visual understanding}. The proposed CFT-CLIP could assist them with aligning the vision and language representation by adopting the counterfactual text. Third, this study could be extended to the broader research on computational social science. As similarly done in previous studies~\cite{10.1145/3524273.3532891,oostdijk-etal-2020-connection}, one could extend the target problem by addressing it as a ranking task rather than binary classification. It could enable a news application that recommends a suitable thumbnail image for a given news article. Additionally, the proposed CFT-CLIP could be adopted for automated fact verification involving image evidence~\cite{luo-etal-2021-newsclippings,mishra2022factify,yao2023end}.

\section*{Limitations}
This study bears several limitations. First, the scale of NewsTT is limited. Since the annotation task is complicated, we chose to do an in-lab annotation to provide high-quality labels for evaluation. Since the dataset reflects the distribution of real-world news articles and the label quality is high, it can be used for a reliable evaluation corpus. Future studies could scale up the dataset via crowdsourcing with the annotation scheme developed in this study. Second, since we continued to pretrain the CLIP encoders, CFT-CLIP inherits the weakness of the CLIP vision encoder. CLIP might be culturally biased toward the Western countries where the pretraining dataset may originate. The pretrained CLIP checkpoint used in this study cannot handle the entire body text as input because the maximum token length is 77, which is smaller than the average body text length. While the use of summary text mitigates the limitation, future studies could use a text encoder that can handle a long sequence to exploit the entire news article. Third, this study focused on developing a zero-shot classifier based on a CLIP-like dual encoder, which does not involve labeled data for training. The performance could be boosted by developing a fine-tuned classifier or few-shot prompt learning methods. 

\section*{Ethics Statement}
This study introduced CFT-CLIP, a contrastive learning framework for training an image-text multimodal encoder. For pretraining, this study used publicly available news articles shared by news media. While we tried to have a high-quality corpus for pretraining, it is possible that the model learned hidden biases in online news. Also, Since CFT-CLIP was updated from the pretrained CLIP weights, it may inherit the bias of CLIP. A user should be cautious about applying the method to problems in a general context and be aware of a potential bias. We have fewer privacy concerns because our study used openly accessible news data that may follow strict internal guidelines according to journalism principles. The NELA-GT-2021 was shared under the license of CC BY-NC 4.0, and the BBC corpus was shared under the MIT license. We will share NewsTT with CC BY NC 4.0. Some of the text was edited using AI assistants, such as ChatGPT and Grammarly.

\section*{Acknowledgements}
Kunwoo Park is the corresponding author. This work was supported by Innovative Human Resource Development for Local Intellectualization program through the Institute of Information \& Communications Technology Planning \& Evaluation (IITP) grant funded by the Korean government (MSIT) (IITP-2024-RS-2022-00156360).

\bibliography{acl_latex}

\appendix
\section*{Appendix}

\setcounter{table}{0}
\renewcommand{\thetable}{A\arabic{table}}
\setcounter{figure}{0}
\renewcommand{\thefigure}{A\arabic{figure}}

\section{Unlabeled data characteristics}
\label{sec:data_analysis}
Table~\ref{abl:data_analysis} shows the data characteristics for the unlabeled train dataset. In reference to Table~\ref{tab:data_analysis}, we found that the labeled data follows a similar distribution to the unlabeled NELA across various dimensions. The labeled summary text tends to have different characteristics from the unlabeled BBC summary dataset because the former was generated by GPT yet the latter is written by the news editor.

\begin{table}[ht]
\small
\centering
\begin{tabular}{cccc}
\toprule
& NELA & \multicolumn{2}{c}{BBC} \\\midrule
& Title & Title & Summary \\\midrule
Words & 14.3 & 9.3 & 25.4 \\
Nouns & 3.24 & 2.56 & 5.22 \\
Verbs & 1.67 & 1.07 & 2.44 \\
Adjectives & 0.94 & 0.61 & 1.54 \\
Named entities & 2.14 & 1.73 & 2.71 \\\bottomrule
\end{tabular}
\caption{Data distribution}
\label{abl:data_analysis}
\end{table}

\section{Configuration details} \label{app:config}

We ran experiments on a machine equipped with AMD Ryzen Threadripper Pro 5975WX CPU, three Nvidia RTX A6000 GPUs (48GB per GPU), and 256GB RAM. We trained the models with the AdamW optimizer with the initial learning rate of 1e-4, updated by the cosine annealing scheduler. The minibatch size is 128. The temperature $\tau$ in the loss equation is 0.05, following \citet{gao2021simcse}. Other hyperparameters were optimized by random search using a validation set. Model training was early-stopped when the validation loss was not decreased five times consecutively, measured for every 20 iterations. The experiments were conducted on Python 3.9, Pytorch 1.10.1, Transformers 4.29.2, LAVIS 1.0.2, and SentenceTransformer 2.2.2. Five random seeds were used for repeated experiments: 0, 1, 2, 3, and 4. The temperature used for adjusting the masked token prediction is set as 2.0. 

The pretrained model checkpoints used for the experiments are as follows:
\begin{itemize}
    \item CLIP: \url{https://huggingface.co/openai/clip-vit-large-patch14}
    \item BLIP: blip\_base (LAVIS)
    \item BLIP-2: blip2\_pretrain (LAVIS)
    \item BLIP-2$+$SBERT: pretrain\_opt2.7b (LAVIS), all-MiniLM-L6-v2 (SentenceTransformer)
    \item Named Entity Recognizer: \url{https://huggingface.co/spacy/en_core_web_trf}
    \item Masked Language Model: \url{https://huggingface.co/bert-base-uncased}
    
\end{itemize}

\section{Ablation experiments}

\paragraph{Reference text} We compared the performance of using news title and summary text as reference text $T$ in the zero-shot classifier. Table~\ref{abl:infer_text} presents the results, showing that using the summary text could achieve a better outcome. Thus, we used the summary text for inference.

\begin{table}[h]
\small
\centering
\begin{tabular}{cccc}
\toprule
$T$ & F1 & Spearman\\\midrule
Summary & \textbf{0.815}$\pm$0.003 & \textbf{0.491}$\pm$0.005 \\
Title & 0.759$\pm$0.003 & 0.386$\pm$0.003\\
\bottomrule
\end{tabular}
\caption{Performance by the reference text type during inference}
\label{abl:infer_text}
\end{table}

\paragraph{Transfer learning} 

To understand whether the proposed contrastive learning framework can lead to a better model than the standard CLIP objective without transfer learning, we conducted experiments that update the parameters of a vision and language bi-encoder from scratch by the CLIP and CFT-CLIP objectives. We used the transformer models with the same architecture but with randomly initialized parameters. We used the BBC dataset with summary text for training, and all parameters were updated until the 20th epoch. The other settings remained the same as those used in the main experiment. Table~\ref{abl:pretraining} shows the results, indicating that CFT-CLIP outperformed CLIP with an f1 of 0.703 and a Spearman coefficient of 0.06. This suggests that contrasting with the counterfactual text can make the vision language bi-encoder learn the cross-modal matching ability for the target task. However, its performance was lower than that of the pretrained CLIP reported in Table~\ref{eval}, which achieved an f1 of 0.763 and a Spearman coefficient of 0.409. Given that the pretrained CLIP was trained on a web-scaled dataset, we guess the performance degradation originated from the scale of the pretraining corpus.

\begin{table}[h]
\small
\centering
\begin{tabular}{cccc}
\toprule
Model & F1 & Spearman\\\midrule
CFT-CLIP    & \textbf{0.703$\pm$0.008} & \textbf{0.060$\pm$0.016} \\
CLIP        & 0.625$\pm$0.005 & 0.045$\pm$0.007 \\\bottomrule
\end{tabular}
\caption{Results without transfer learning}
\label{abl:pretraining}
\end{table}

\section{Data examples}

\subsection{Failed prediction}

\begin{figure*}[ht]
    \centering
    \includegraphics[width=0.7\linewidth]{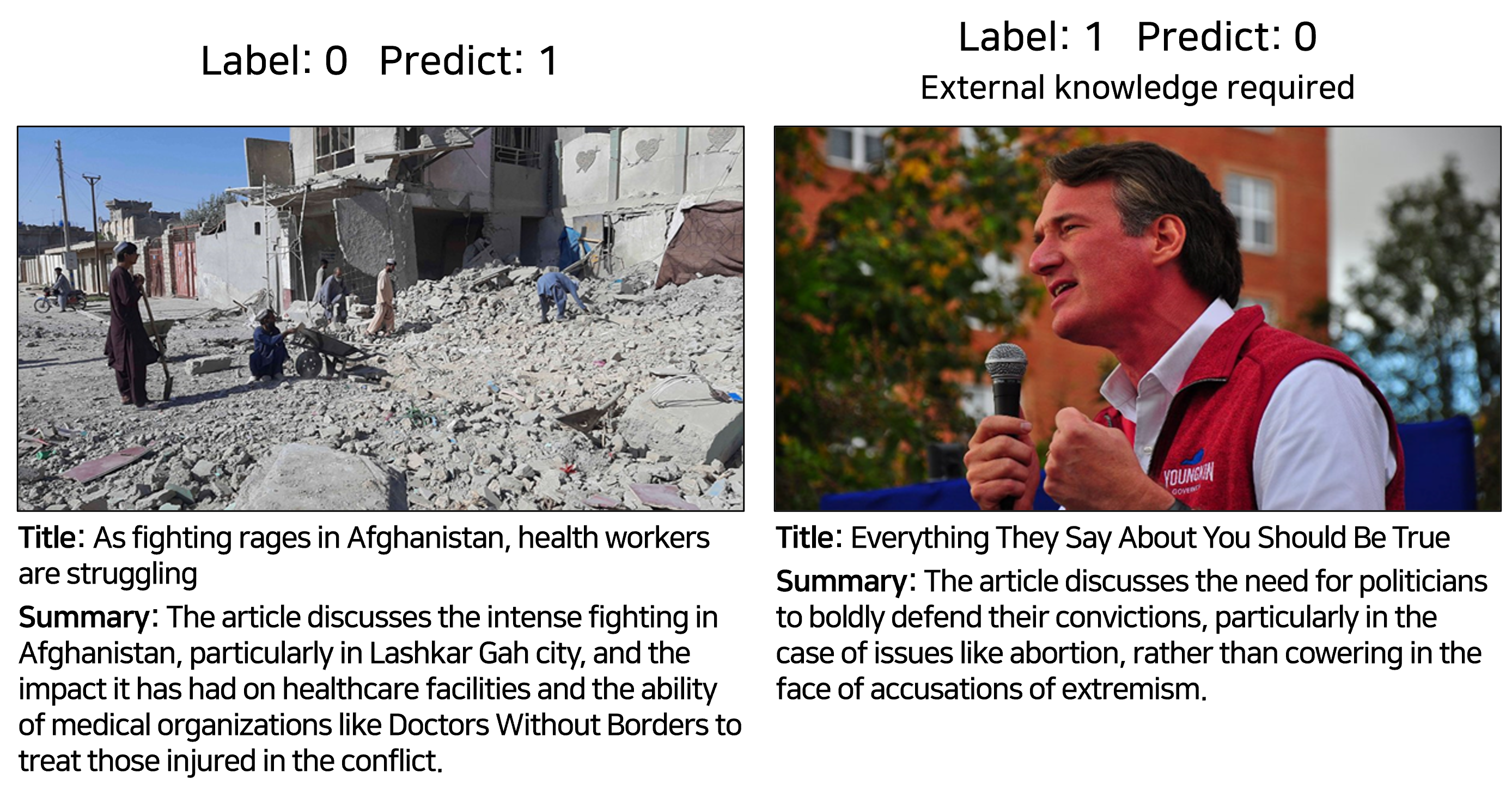}
    \caption{More error examples}
    \label{fig:qualitative_failed_example_add}
\end{figure*}

Table~\ref{fig:qualitative_failed_example_add} presents two examples where the CFT-CLIP model made a failed prediction. The first example presents the false positive case where the model returns a high similarity score for the unrepresentative thumbnail. The example on the right shows an error case where referring to external knowledge is required. The person in the image is Glenn Youngkin, the governor of Virginia in the US. A model could not return a high similarity score without knowing his background.

\subsection{Counterfactual text}
\label{sec:app:counter}

Table~\ref{tab:counterfactual_examples} presents several examples derived from our counterfactual text generation method. Overall, the proposed method successfully generates the counterfactual text by selecting appropriate entities and substituting them with other entities. On the other hand, there are several cases where the generated token breaks the grammar. For instance, in the fourth example, the term `Oisin Murphy' was replaced by `offensive.' Such anomalies may arise from the sampling process employed, which aims to prevent the recreation of the original text. Despite its imperfect structure, the generated sentence can still serve as a hard negative sample during contrastive update, given that its general context is preserved.

\begin{table*}[t]
    \small
    \centering
    \begin{tabular}{cc}
    \toprule
    Original text & Generated text \\\midrule
    \parbox[t]{7cm}{Op-Ed: Memo to Saddleback Church: Replacing Pastor \ul{Rick Warren} is a minefield} & \parbox[t]{7cm}{Op-Ed: Memo to Saddleback Church: Replacing Pastor \ul{Parker} is a minefield} \\
    \parbox[t]{7cm}{‘The greatest striker’: \ul{Gerd M\"uller}, legendary German forward, dies aged 75} & \parbox[t]{7cm}{‘The greatest striker’: \ul{Joseph}, legendary German forward, dies aged 75}\\
    \parbox[t]{7cm}{\ul{Matt Gaetz} and wingman facing 'mutually assured destruction' after confession letter: legal expert} & \parbox[t]{7cm}{\ul{Pennant} and wingman facing 'mutually assured destruction' after confession letter: legal expert} \\
    \parbox[t]{7cm}{\ul{William Buick} treble sets up Flat jockeys’ title race for dramatic finish as gap closes on \ul{Oisin Murphy}} & \parbox[t]{7cm}{\ul{Davidson} treble sets up Flat jockeys’ title race for dramatic finish as gap closes on \ul{offensive}}\\
    \parbox[t]{7cm}{\ul{Michael Douglas} says it was ‘uncomfortable’ for him and Catherine Zeta-Jones to share Mallorcan home with his ex} & \parbox[t]{7cm}{\ul{Novella} says it was ‘uncomfortable’ for him and Catherine Zeta-Jones to share Mallorcan home with his ex}\\
    \parbox[t]{7cm}{Experts say  \ul{Jussie Smollett} is in ‘matrix of arrogance’ as he awaits sentencing} & \parbox[t]{7cm}{Experts say \ul{Toni} is in ‘matrix of arrogance’ as he awaits sentencing}\\
    \parbox[t]{7cm}{\ul{Ted Nugent} tests positive for coronavirus after calling pandemic a ‘scam’} & \parbox[t]{7cm}{\ul{Danny} tests positive for coronavirus after calling pandemic a ‘scam’}\\
    \parbox[t]{7cm}{\ul{Jennifer Aniston} Explained How Therapy Helps Her Deal With The "Tough Stuff" Of Being Famous} & \parbox[t]{7cm}{\ul{Charles} Explained How Therapy Helps Her Deal With The "Tough Stuff" Of Being Famous}\\
    \parbox[t]{7cm}{Thousands mark anniversary of Kremlin critic \ul{Nemtsov}’s murder} & \parbox[t]{7cm}{Thousands mark anniversary of Kremlin critic \ul{pastor}’s murder}\\
    \parbox[t]{7cm}{Nets Disregard AG \ul{Garland} Grilled in Hearing for Targeting Parents} & \parbox[t]{7cm}{Nets Disregard AG \ul{Cicero} Grilled in Hearing for Targeting Parents}\\
    \parbox[t]{7cm}{\ul{Travis Kelce} Is Borderline Unrecognizable Without Facial Hair} & \parbox[t]{7cm}{\ul{Annie} Is Borderline Unrecognizable Without Facial Hair}\\
    \parbox[t]{7cm}{Billionaire \ul{Ken Griffin} bought a copy of the US Constitution for \$43.2m because his son asked him to} & \parbox[t]{7cm}{Billionaire \ul{Dublin} bought a copy of the US Constitution for \$43.2m because his son asked him to}\\
    \parbox[t]{7cm}{Desperate \ul{Chuck Todd} Hopes \ul{Trump} Will Deflect Media ‘Spotlight’ From Dem ‘Problems’} & \parbox[t]{7cm}{Desperate \ul{Joe} Hopes \ul{Radha} Will Deflect Media ‘Spotlight’ From Dem ‘Problems’}\\
    \parbox[t]{7cm}{\ul{Chris Cuomo's} Book Contract Dropped By HarperCollins} & \parbox[t]{7cm}{\ul{Harriet} Book Contract Dropped By HarperCollins}\\
    \parbox[t]{7cm}{\ul{Jen Psaki} shoots down a reporter comparing \ul{Biden} to his predecessor: \ul{Trump} suggested ‘people inject bleach’} & \parbox[t]{7cm}{\ul{person} shoots down a reporter comparing \ul{virus} to his predecessor: \ul{Walden} suggested ‘people inject bleach’}\\
    \parbox[t]{7cm}{What’s the Deal With \ul{Gavin Newsom}? 5 Plausible Theories To Explain His Mysterious Hiatus} & \parbox[t]{7cm}{What’s the Deal With \ul{diplomacy}? 5 Plausible Theories To Explain His Mysterious Hiatus}
    \\\bottomrule 
    \end{tabular}
    \caption{Counterfactual text generation examples}
    \label{tab:counterfactual_examples}
\end{table*}

\section{API usage details}
\label{appendix:chatgpt}

We used OpenAI API to use GPT 3.5-Turbo. We obtained news summaries for the labeled dataset and generated counterfactual text for the summary text of the BBC unlabeled pretraining corpus. In total, the API call cost \$13.33. Below are the prompts used in the experiments.

\paragraph{News summarization} 

\begin{quote}
\small
Article:\{text\}

Summarize the article in one sentence.
\end{quote}

\paragraph{Counterfactual text generation}

\begin{quote}
\small
Create a counterfactual news summary by modifying the actors of news events:\{text\}

Answer in JSON. The JSON should be a string of dictionaries whose keys are "counterfactual".
\end{quote}

\section{Annotation details and guidelines}
\label{sec:guideline}

We hired two male and one female student from Soongsil University. The annotators were trained by using the guidelines in Table~\ref{tab:guideline}. The original guideline was in another language, and we present its English-translated version. All the annotators were paid \$0.1 per example.

\begin{table*}[ht]
\centering
\resizebox{.99\linewidth}{!}{%
\begin{tabular}{|p{16cm}|}
\hline
\textbf{Task overview:}\\
In this task, you are asked to answer whether a given news image represents the actors of the news events.\\

\\
\textbf{Instruction:}   \\
- Q1. Identify news actors in the text. The actors can be expressed as named entities, proper nouns, or common nouns.\\
- Q2. Does the image display the news actors identified in Q1?\\
- Q3. Identify the visually presented news actors in the image.\\

\\
\\
\textbf{Examples}: \\
\\

\includegraphics[width=1\textwidth]{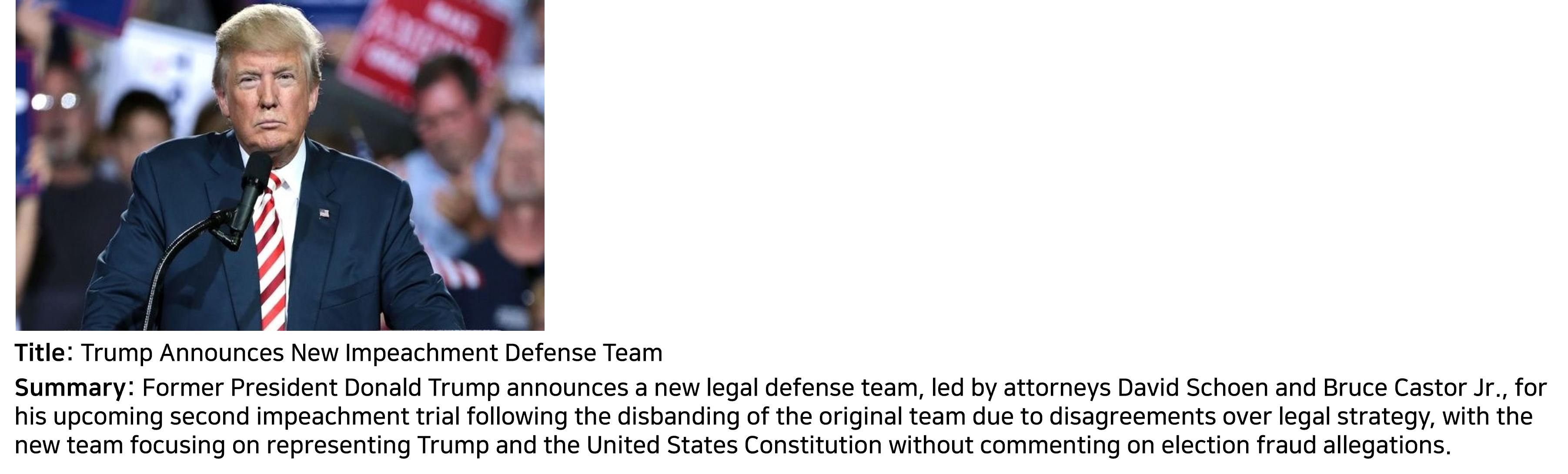} \\

- Q1: Trump, Impeachment Defense Team, David Schoen, Bruce Castor Jr., impeachment trial, disagreements, legal strategy, United States Constitution, election fraud allegations \\
- Q2: Y \\
- Q3: Trump

\\
(More examples) \\\hline
\end{tabular}%
}
\caption{Translated annotation guideline}
\label{tab:guideline}
\end{table*}

\end{document}